\documentclass[letterpaper]{article} 
\usepackage{aaai2026}  
\usepackage{times}  
\usepackage{helvet}  
\usepackage{courier}  
\usepackage[hyphens]{url}  
\usepackage{graphicx} 
\usepackage{natbib}  
\usepackage{caption} 
\usepackage{algorithm}
\usepackage{algorithmic}

\usepackage{newfloat}
\usepackage{listings}
\DeclareCaptionStyle{ruled}{labelfont=normalfont,labelsep=colon,strut=off} 
\floatstyle{ruled}
\newfloat{listing}{tb}{lst}{}
\floatname{listing}{Listing}
\usepackage{bm}
\usepackage{amsmath} 
\usepackage{booktabs}
\usepackage{multirow}

\title{Semi-supervised Latent Disentangled Diffusion Model for Textile Pattern Generation}
\author{
    Chenggong Hu\textsuperscript{\rm 1}\equalcontrib, Yi Wang\textsuperscript{\rm 1}\equalcontrib, Mengqi Xue\textsuperscript{\rm 2}, Haofei Zhang\textsuperscript{\rm 1}, Jie Song\textsuperscript{\rm 1}, Li Sun\textsuperscript{\rm 1}\footnote{Corresponding author.}
}
\affiliations{
    \textsuperscript{\rm 1}Zhejiang University, \textsuperscript{\rm 2}Hangzhou City University \\ 
    
    {\{huchenggong, y\_w, sjie,  haofeizhang, lsun\}@zju.edu.cn, mqxue@hzcu.edu.cn}, 
}

\usepackage{bibentry}

\begin{document}

\maketitle

\begin{abstract}
Textile pattern generation (TPG) aims to synthesize fine-grained textile pattern images based on given clothing images. Although previous studies have not explicitly investigated TPG, existing image-to-image models appear to be natural candidates for this task. However, when applied directly, these methods often produce unfaithful results, failing to preserve fine-grained details due to feature confusion between complex textile patterns and the inherent non-rigid texture distortions in clothing images. In this paper, we propose a novel method, SLDDM-TPG, for faithful and high-fidelity TPG. Our method consists of two stages: (1) a latent disentangled network (LDN) that resolves feature confusion in clothing representations and constructs a multi-dimensional, independent clothing feature space; and (2) a semi-supervised latent diffusion model (S-LDM), which receives guidance signals from LDN and generates faithful results through semi-supervised diffusion training, combined with our designed fine-grained alignment strategy. Extensive evaluations show that SLDDM-TPG reduces FID by $4.1$ and improves SSIM by up to $0.116$ on our CTP-HD dataset, and also demonstrate good generalization on the VITON-HD dataset. Our code is available \url{https://github.com/Cg-Hu/SLDDM-TPG}.
\end{abstract}


\section{Introduction}
Contemporary image-to-image diffusion models demonstrate remarkable generative capabilities \cite{zhang2024ssr,li2024blip,wang2024msdiffusion,qi2024deadiff,ju2024brushnet} across many tasks, such as virtual try-on \cite{choi2024improving} and scene generation \cite{huang2025midi}. However, in the inherently visual and artistic field of fashion design, the adoption of such powerful generative tools for creating fashion designs remains limited. Progress in this area could substantially lower design costs while preserving creative fidelity. In this work, we refer to this relatively underexplored area as \textit{Textile Pattern Generation} (TPG), a design process that integrates varied shapes, colors, and textures to produce novel and visually appealing patterns. TPG is central to the textile sector, with wide-ranging influence on fashion design, apparel creation, and textile manufacturing.

\begin{figure}
    \centering
    \includegraphics[width=0.88\linewidth]{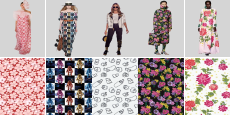}
    \caption{Textile pattern image results generated by our SLDDM-TPG from real worn clothing on our CTP-HD dataset. The first row show the clothing images and the second row show the generated textile pattern images.}
    \label{fig:firstrow}
    \vspace{-18pt}
\end{figure}

The core goal of TPG is to extract the complex textile pattern contents and design from a natural clothing image and reconstruct them into a highly realistic textile pattern image (simply referred to as the pattern image), as illustrated in Figure~\ref{fig:firstrow}. The synthesized pattern image is expected to meet the following desiderata: (1) the generation should faithfully retain clear and consistent visual details from the clothing image, including not only low-level features such as color and texture, but also complex elements like floral motifs, geometric patterns and so on; (2) the results should exhibit a well-structured overall layout, similar to real pattern images; (3) the outputs must be free from any texture defects, such as \textit{deformations}, \textit{blurriness}, and \textit{occlusions}. However, clothing images often suffer from texture defects, which contrast sharply with well-structured textile pattern images. Therefore, it is essential to eliminate these defects and accurately extract the underlying textile pattern content and design embedded in clothing images. This process naturally leads to feature disentanglement, which is key to enabling faithful and fine-grained reference-based textile pattern generation.

Recently, popular image-to-image (I2I) diffusion models \cite{zhang2024ssr,gou2023taming,gao2024styleshot} appear to be a natural choice for TPG. However, they rarely consider feature disentanglement required for processing clothing images, often leading to unfaithful generations. Specifically, I2I diffusion models typically input a whole image as a reference into the feature encoder and then use its output, the entangled features, to guide the generation. In TPG scenarios, clothing images contain not only the intended pattern content but also unwanted texture defects. Conditioning generation directly on entangled features increases task complexity for diffusion models, making it harder to isolate informative signals and thus impeding the generation of high-fidelity pattern images with fine-grained details. Although text-guided approaches \cite{zhang2024ssr} have been proposed to localize semantic features for disentanglement, the inherent visual complexity of textile pattern designs makes them difficult to describe precisely using text. Moreover, due to the limited availability of training data specific to TPG, these fully supervised learning methods often suffer from overfitting and exhibit poor generalization performance.

To address these challenges in TPG, we propose a semi-supervised latent disentangled diffusion model (SLDDM-TPG), a two-stage framework for generating faithful pattern images from clothing images. In the first stage, to address the unfaithful generation caused by feature confusion in clothing images, we propose a latent disentangled network (LDN) to disentangle useful and irrelevant information. LDN first extracts the \textbf{shared semantic content} between clothing and pattern images using a similarity contrast module (SCM), which serves as generation guidance. Meanwhile, a reverse attention module (RAM) captures \textbf{texture defects} specific to clothing images, which are used as negative prompts. However, clothing images often lack the well-structured layout specific to pattern images. We define this \textbf{structured information} as \textit{flatness}, \textit{clarity}, and \textit{full visibility}. To recover such features, we introduce structure affine transformations (SATs), which map low-level defect features to structured representations. SATs are jointly trained with RAM to ensure feature stability and prevent drift from the original latent space. In the second generation stage, we propose a semi-supervised latent diffusion model (S-LDM) that strategically leverages labeled and additional unlabeled data to improve generation quality and generalization, guided by LDN features. The framework adopts a cascaded two-stage architecture consisting of a denoising process and an alignment process. The denoising process learns the distribution of pattern images through denoising training. The alignment process treats the model-predicted image as a semi-supervised optimization target, encouraging the output to align not only with the general data distribution but also with the specific content of each clothing image, guided by our proposed stable transformation domain (STD) loss. Recognizing local similarity as an intrinsic property of pattern images, we propose a convolutional local similarity (CLS) module to enforce local consistency via localized feature matching. We also present CTP-HD, a high-resolution dataset of paired clothing and pattern images for training.

In summary, our main contributions are as follows: (1) we propose SLDDM-TPG, the first model to achieve faithful and fine-grained clothing-based pattern generation; (2) we introduce CTP-HD, the first high-resolution paired dataset bridging clothing and pattern images; (3) our proposed LDN addresses the feature confusion in clothing images by disentangling their representations into multi-dimensional feature spaces; (4) we propose S-LDM that enables to utilize unlabeled data and we design an alignment process with a novel STD loss and CLS module to refine generation quality.

\section{Related Work}
\noindent\textbf{Textile Pattern Generation.} There are existing methods similar to TPG \cite{wu2024application,pang2019nad}, but they all focus on unconditional synthesis without clothing image guidance. As a result, they are inherently incapable of generating garment-specific patterns from real-world clothing images, often producing uncontrollable outputs with low fidelity. Most GAN-based approaches \cite{fayyaz2020textile, wang2021dyeing} additionally suffer from mode collapse and limited pattern diversity. Meanwhile, 3D-to-2D methods \cite{lu2017new, choi2007method} rely on expensive 3D scanning data, which significantly limits the practical applicability. In contrast, our framework, SLDDM-TPG, enables controllable generation of real-world textile pattern images conditioned on RGB clothing images through a latent diffusion model. It achieve precise content alignment with the reference clothing image while preserving the well-structured overall layout of the real patterns.

\noindent\textbf{Image-to-Image Diffusion Models.} Recent advancements in diffusion models have significantly enhanced image-to-image tasks, including image inpainting, reference-based generation, and style transfer. For image inpainting, StrDiffusion~\cite{Liu_2024_CVPR} uses structure-guided denoising to preserve semantic consistency and DCI-VTON~\cite{gou2023taming} treats virtual try-on as structure-aware inpainting guided by parsing maps. In reference-based generation, SSR-Encoder~\cite{zhang2024ssr} encodes image subject across multiple scales, IP-Adapter~\cite{ye2023ip} injects image features into diffusion models via cross-attention for subject-driven synthesis, and UniCon~\cite{li2024unicon} models joint distributions over image pairs for direct reference-guided generation. For style transfer, StyleShot~\cite{gao2024styleshot} introduces a style-aware encoder to extract style embeddings across diverse styles, while OSASIS~\cite{cho2024one} leverages Diff-AE~\cite{preechakul2022diffusion} for one-shot facial style transfer. However, these methods overlook feature confusion in clothing images, resulting in unfaithful outputs. In contrast, our LDN addresses this issue by disentangling garment features for more faithful generation.

\section{Method}
\begin{figure*}[!t]
    \centering
    \includegraphics[width=1\textwidth, trim=0 0 30 1, clip]{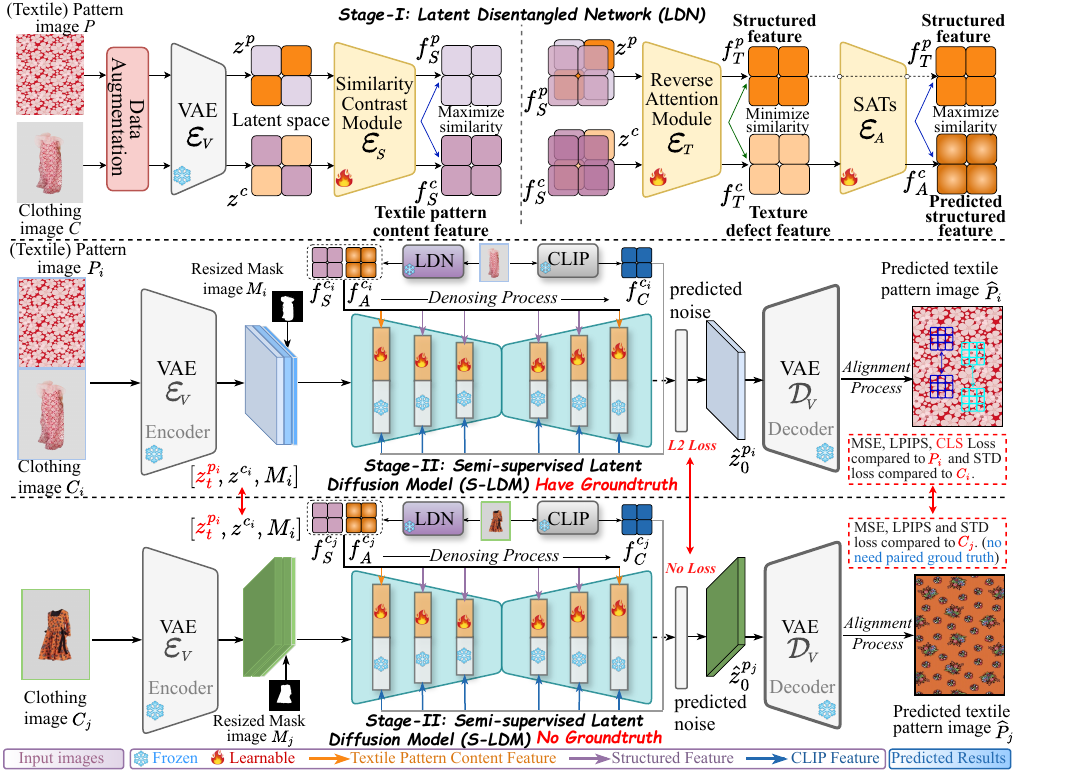}
    \caption{The framework of SLDDM-TPG. During training, LDN is first trained to disentangle features of clothing images ${C}$ to \textbf{textile pattern content feature} ${f_S^c}$, predicted \textbf{structured feature} $f_A^c$, and \textbf{texture defect feature} $f_T^c$. Then S-LDM is trained, guided by LDN's features to generate pattern images ${P}$ aligned with image ${C}$. There are data with and without ground truth in a batch. The two S-LDM share the same network and parameters, but their inputs, calculated losses, and comparison objects of losses are different, as can be seen in red parts in \textit{stage-II}. During inference, a clothing image ${C}$ is fed into LDN, whose output features are passed to S-LDM as conditions, and $T$-step denoising is then applied to the noise input to generate the final result.}
    \label{fig:framework}
\end{figure*}

Our method targets faithful and fine-grained generation of pattern images from clothing images. We adopt Stable Diffusion V1-5 \cite{rombach2022high} as the backbone and introduce an adapter \cite{ye2023ip} to incorporate guidance. As shown in Figure \ref{fig:framework}, our method consists of two stages: (1) a latent disentangled network (LDN), employing three collaborative modules to resolve feature confusion and extract diverse features from clothing images; (2) a semi-supervised latent diffusion model (S-LDM), guided by the extracted features and an alignment process, and trained with unlabeled and limited labeled data to produce faithful results.

\subsection{Latent Disentangled Network}
As shown in Figure~\ref{fig:framework}, our three-module latent disentangled network (LDN) effectively addresses feature confusion in clothing representations. Specifically, given a clothing image $C$, LDN disentangles the textile pattern content feature $f_S^c$ shared with the pattern image $P$, extracts the clothing-specific texture defect feature $f_T^c$, and predicts the structured feature $f_A^c$, which is absent in the clothing image $C$ but important for faithful reconstruction.

\label{sec:LDN}

\noindent\textbf{Similarity Contrast Module.} It is evident that clothing images $C$ and pattern images $P$ share semantic content that requires targeted reconstruction. To extract this shared information from image $C$, we introduce a similarity contrast module (SCM), which adopts SimSiam \cite{chen2021exploring} to perform contrastive learning between clothing images $C$ and pattern images $P$ for extracting the shared textile pattern content feature $f_S^c$. SimSiam applies different data augmentations to a sample to produce two views, analogous to a pair of clothing image $C$ and pattern image $P$, for contrastive learning, as shown in Figure \ref{fig:framework}. SimSiam supports small-batch training and does not rely on negative samples, achieving strong performance via a stop-gradient operation. In our setting, training is conducted in the latent space, and we modify the original augmentation strategies to better suit our clothing and pattern images,  thereby improving generalization performance. The optimization objective is defined as follows:
\begin{equation}
    \mathcal{L}_{SCM} = -\frac{1}{2}(\text{sim}(s_{\mathcal{P}}^p, \mathrm{sg}(s_{\mathcal{E}}^c)) +  \text{sim}(s_{\mathcal{P}}^c, \mathrm{sg}(s_{\mathcal{E}}^p))),
\label{equ:simsiam}
\end{equation}
where $\text{sim}(\cdot,\cdot)$ denotes cosine similarity and $\text{sg}(\cdot)$ represents the stop-gradient operation. Here, $s_{\mathcal{E}}^{c}$ and $s_{\mathcal{E}}^{p}$ denote the encoder outputs of SimSiam for the clothing and pattern images, respectively, while $s_{\mathcal{P}}^{c}$ and $s_{\mathcal{P}}^{p}$ denote the corresponding predictor outputs. Like SimSiam, we adopt the encoder output as the textile pattern content feature $f_S^c$ after training.



\noindent\textbf{Reverse Attention Module with SATs.} 
Besides the shared content feature $f_S^c$, the clothing image $C$ contains texture defects and lacks the structured feature specific to $P$. To avoid interference caused by these defects, we introduce the reverse attention module (RAM) to extract them as negative prompts. To train RAM, we first freeze the SCM. For $C$, we use $f_S^c$ as the query ($Q$), and the latent feature $z^c$ as both the key ($K$) and value ($V$). The original $N$ cross-attention layers use $Q$ to query semantically similar features from $K$, and compute attention weights $A_i$ to aggregate the corresponding values $V$ \cite{zhang2024ssr}, as defined below:
\begin{equation}
A_{i}=\text{Softmax} \left( \frac{Q_iK_i^\top}{\sqrt{d}} \right),\quad i\in[1,N].
\label{equ:triplet}
\end{equation}
However, in RAM, we reverse $A_i$ as $A^r_i=\text{Normalize}(1-A_i)$ in each layer to amplify coarse dissimilarities between $Q$ and $K$. We then compute the new weighted sum and concatenate it with the latent feature $z^c$ as a residual, forming the initial texture defect feature $f^c_T$, as follows:
\begin{equation}
\text{Initial: }f^c_T=z^c+\sum A^r_{i}\cdot V_i,\quad i\in[1,N].
\label{equ:triplet}
\end{equation}
The same operation is applied to $P$ to obtain the initial structured feature $f_T^p$. We then train the model to push $f_T^c$ and $f_T^p$ apart in the feature space, as they exhibit inherently opposite characteristics. However, the single constraint often leads to feature drift beyond the original latent space, resulting in the loss of basic semantics (details are provided in our ablation study). To ensure stable convergence during RAM training and address the absence of structured features in clothing images, we leverage the observation that both texture defect feature $f^c_T$ and structured feature $f^p_T$ are both low-level representations, but exhibit contrasting properties. This motivates us to apply affine transformations to convert $f_T^c$ into $f_T^p$. To this end, we propose structured affine transformations (SATs) and jointly train them with RAM. SATs consist of a set of learnable affine transformation networks. For example, we design a convolutional filtering network for sharpness-adjusting transformations (details are provided in the Appendix). Inspired by \cite{schroff2015facenet}, we design a texture triplet loss to enforce separation between the texture defect feature $f_T^c$ and the structured feature $f_T^p$, while encouraging $f_T^c$ to be transformable toward $f_T^p$. The loss is defined as:
\begin{equation}
\mathcal{L}_{TRIPLET}=\|f^c_A-f^p_T\|^2_2-\|f^c_T-f^p_T\|^2_2+\alpha,
\label{equ:triplet}
\end{equation}
where $f^c_A$ is the structured feature predicted by SATs from $f^c_T$ of $C$, and $\alpha$ is a margin that is enforced between $f^c_T$ and $f^p_T$ pair. In summary, the loss of LDN is as follows:
\begin{equation}
\mathcal{L}_{LDN}=\mathcal{L}_{SCM}+\mathcal{L}_{TRIPLET}.
\label{equ:ldn}
\end{equation}

Details of the data augmentations and network structures for the three modules are provided in the Appendix.

\subsection{Semi-supervised Latent Diffusion Model}
\label{sec:S-LDM}
Our S-LDM includes denoising distribution learning using labeled data, and an alignment process that enables semi-supervised~\cite{yin2025self,he2024progressive} training by leveraging unlabeled data through similarity constraints with reference images.


\noindent\textbf{Denoising Distribution Learning Process.} As shown in Figure~\ref{fig:framework}, we freeze the backbone and use the features of the frozen LDN to train an added adapter. Inspired by \cite{voynov2023p+}, the different layers of cross attention in the denoising UNet \cite{ronneberger2015u} are responsible for synthesizing different aspects of content. Thus, we feed the high-resolution coarse layer of the UNet with the low-level structured feature $f_A^c$ and the low-resolution fine layer with the high-level textile pattern content feature $f_S^c$ to generate different parts of the pattern images. We then use the texture defect feature $f_T^c$ as a negative prompt, and perform conditional generation through CFG \cite{ho2022classifier}. During training, S-LDM (denoted as $\bm\epsilon_\theta$) requires the $t$-step noised latent feature $z_t^p$ of ground truth $P$. We then concatenate it with the latent feature $z^c$ of clothing image $C$ and its resized mask $M$ to form the input $\bm\phi_t^p = [z_t^p, z^c, M]$ as shown in Figure~\ref{fig:framework}, where $[\cdot,\cdot,\cdot]$ denotes channel-wise concatenation. The denoising process is trained to learn the distribution of the general pattern images $P$ as follows:
\begin{equation}
\label{equ:dp}
    \mathcal{L}_{DP}=\|{\bm\epsilon}_\theta(\bm{\phi}^p_t,f^{c}_S,f^c_A,f^c_T,f^c_C, t)-\bm{\epsilon}\|^2_2,
\end{equation}
where $f^c_C$ is the feature of the frozen CLIP \cite{radford2021learning} image encoder, and $\bm\epsilon$ is the actual $t$-step noise added to the latent feature $z^p$ of the ground truth image $P$.

\noindent\textbf{Alignment Process.} This process enables generation alignment and supports semi-supervised learning. Specifically, by our designed module and loss functions that constrain the predicted output $\hat{P}$ to be similar to either the ground truth $P$ (if available) or the input clothing image $C$, we achieve supervision even without annotations. In this way, enforcing alignment with $C$ allows the model to learn from unlabeled data, as illustrated by the \textit{red arrows and dashed boxes} in Figure~\ref{fig:framework}. The predicted $\hat{P}$ is computed as follows: 
\begin{equation}
\label{eq:onestep}
    \hat{P}=\mathcal{D}_V(\frac{{z}^c_{t}-\sqrt{1-{\bar\alpha}_{t}}\cdot\bm\epsilon_{\theta}(\bm{\phi}^*_t,f_{S}^c,f_A^c,f_T^c,f_C^c, t)}{\sqrt{{\bar\alpha}_{t}}}),
\end{equation}
where $\bm{\phi}^*_t$ can be either  $\bm{\phi}^p_t$ or $\bm{\phi}^c_t=[z_t^c, z^c, M]$ (no ground truth). $z_t^c$ is the t-step noised
latent feature of $C$. The details of our designing module and loss functions are as follows:

\noindent\textit{(1) Stable Transformation Domain (STD) Loss.} To ensure stable transformation from a clothing image $C$ to the generated pattern $\hat{P}$ during semi-supervised training, we propose a STD loss that jointly aligns cross-sample transformation behavior and constrains intra-domain consistency. The first part of the loss compares the distributional transformation vector $(\mathbf{v}_{pred} = f_S^{\hat{p}} - f_S^c)$ from $C$ to $\hat{P}$ with a reference transformation $(\mathbf{v}_{real} = f_S^{p_r} - f_S^{c_r})$ obtained from \textbf{randomly sampled} real paired data $C_r$ and $P_r$ in each mini-batch, encouraging the model to mimic realistic shifts in content space even in the absence of ground truth supervision. The second part further promotes structural stability by enforcing similarity between the variations within each domain—comparing $C$ and $C_r$ $(\mathbf{d}_{ref} = f_S^c - f_S^{c_r})$ on the one hand, and $\hat{P}$ and $P_r$ $(\mathbf{d}_{gen} = f_S^{\hat{p}} - f_S^{p_r})$ on the other. This dual objective prevents overfitting to synthetic transformations and encourages reliable structure retention. Here, $f^*_S$ denotes the textile pattern content feature. The full transformation loss is:
\begin{equation}
\label{eq:mbd}
\mathcal{L}_{STD}=\|\mathbf{v}_{real}-\mathbf{v}_{pred}\|_2^2 + \|\mathbf{d}_{gen}-\mathbf{d}_{ref}\|_2^2
\end{equation}
Together, these constraints stabilize the semi-supervised learning process and enable consistent textile pattern generation, even in the absence of paired data supervision. 

\noindent\textit{(2) Convolutional Local Similarity (CLS) Module.} We observe that pattern images exhibit content periodicity (local similarity), characterized by repeating contents, textures and structures. Based on this, we propose a CLS module through local region similarity maps alignment to enhance the self-similarity of the generation. Specifically, we randomly select $N$ region matrices of size $64\times64$ from a real pattern image ${P}$ as kernels $\mathcal{K}_i$, where $i\in[1,N]$, and compute the cosine similarity with other same size regions of itself to produce $N$ local similarity maps, as shown in Figure \ref{fig:framework}. Then, entries in the $N$ maps that exceed the threshold $\mathcal{T}$ are set to $1$, indicating regional similarity, others are set to $0$, indicating dissimilarity, yielding $N$ binary local similarity maps $W_i^p$ as supervision labels. Local similarity maps $W_i^{\hat{p}}$ of the predicted $\hat{P}$ are obtained in the same way. To evaluate whether $P$ and $\hat{P}$ share similar local structures, we compare them using the Dice coefficient \cite{milletari2016v}:
\begin{equation}
\label{eq:dice}
\mathcal{L}_{CLS}=-\sum_{i=1}^N\frac{2\sum(W^{\hat{p}}_i W^p_i)}{{\sum W^{\hat{p}}_i}+\sum W^{p}_i},
\end{equation}
where $\sum$ in the fraction denotes the element-wise sum over each map. We set $N=4$ and $\mathcal{T}=0.7$.

\noindent\textit{(3) Perceptual and Pixel-Level Consistency.} We include LPIPS loss~\cite{dosovitskiy2016generating} and MSE loss to ensure the perceptual and pixel-level similarity between $\hat{P}$ and the ground truth $P$ (if available) or the clothing image $C$. The loss function of the alignment process is defined as:
\begin{equation}
\label{eq:rb_loss}
\mathcal{L}_{AP}=\lambda_1\mathcal{L}_{STD}+\lambda_2\mathcal{L}_{CLS}+\lambda_3\mathcal{L}_{LPIPS}+\lambda_4\mathcal{L}_{MSE}.
\end{equation}
Here, we set $\lambda_1=\lambda_2=1e^{-4}, \lambda_3=1e^{-2}, \lambda_4=1e^{-1}$. The total S-LDM training loss is defined as follows:
\begin{equation}
\label{eq:dpa-dm_loss}
\mathcal{L}_{S-LDM}=\mathcal{L}_{DP}+\mathcal{L}_{AP}.
\end{equation}
The S-LDM preserves the distributional stability of the pattern images and ensures content alignment with the clothing images. The details of MSE and LPIPS losses are provided in the Appendix.

\section{Experiments}
\subsection{Experimental Settings}
\noindent\textbf{Datasets.} 
We introduce a novel high-resolution paired dataset of clothing and textile pattern images for TPG, named the clothing textile pattern dataset (CTP-HD). It contains $9,804$ annotated pairs (clothing images with corresponding pattern ground truths) and over $10,000$ unannotated clothing images. All images are of uniform resolution $501\times821$ pixels, making it the first large-scale dataset for TPG. We also use the virtual try-on dataset, VITON-HD \cite{choi2021viton}, for generalization testing only.

\noindent\textbf{Evaluation Metrics.} We use Learned Perceptual Image Patch Similarity (LPIPS) \cite{dosovitskiy2016generating}, Structural Similarity Index (SSIM) \cite{wang2004image}, and Fréchet Inception Distance (FID) \cite{heusel2017gans} to evaluate visual quality of generated images. In addition, we introduce a new metric, Fourier Periodic Similarity (FPS), to measure content periodicity based on local similarity, as detailed in the Appendix. For generalization testing without ground truth, we use LPIPS and MSE to evaluate visual similarity (VLS), and compute content similarity (CTS) using the textile pattern content feature $f_S$ extracted from SCM.

\noindent\textbf{Baseline Methods.} We compare our model with representative image-to-image generation methods relevant to TPG, including image inpainting (DCI-VTON \cite{gou2023taming}, Paint-by-Example \cite{yang2023paint}, StrDiffusion \cite{Liu_2024_CVPR}), reference-based generation (SSR-Encoder \cite{zhang2024ssr}, IP-Adapter \cite{ye2023ip}), UniCon \cite{li2024unicon} and style transfer (StyleShot \cite{gao2024styleshot}, OSASIS \cite{cho2024one}). Details on the CTP-HD dataset, implementation, and baseline methods can respectively be found in the Appendix.

\subsection{Generation on CTP-HD Dataset}
\begin{figure*}[t]
\centering
    \includegraphics[width=0.99\textwidth, trim=0 0 0 6, clip]{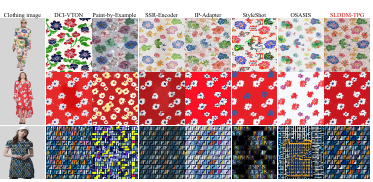}
\caption{Qualitative comparisons of different methods on TPG on our \textit{CTP-HD} dataset \textit{with and without GT}.}
\label{fig:ctp}
\end{figure*}

\noindent\textbf{Qualitative Comparisons.} Figure \ref{fig:ctp} shows the qualitative results of different methods. Image inpainting methods often result in color distortion and misaligned pattern contours. Reference-based generation methods yield coarse, structurally inconsistent patterns due to inaccurate semantic extraction. Style transfer methods often introduces style-color artifacts (e.g. foreground–background hue swapping) and suffers from pattern collapses. By contrast, our method restores fine-grained details and faithfully reconstructs clothing appearance and well-structured overall layout. Besides, we provide more visualization results in the Appendix.

\noindent\textbf{Quantitative Comparisons.} Table~\ref{tab:ctp} presents the quantitative results of different methods. Style transfer methods perform poorly as they fail to capture complex texture content. In addition, the significant domain gap between pattern and natural images makes modeling especially difficult under limited labeled data. Image inpainting methods enable content reconstruction but still fail to deliver satisfactory results due to feature confusion of the clothing images. Reference-based generation methods also struggle to capture complex clothing content when relying on text-driven subject extraction, image encoding, or joint distribution modeling, often resulting in inaccurate features and coarse outputs. In contrast, our method disentangles clothing features into more independent and accurate feature spaces, and improves generation quality through the alignment process, consistently outperforming baselines across all metrics.

\subsection{Generalization Testing on VITON-HD Dataset}
\begin{figure}[!t]
\centering
    \includegraphics[width=0.48\textwidth, trim=0 0 0 5.5, clip]{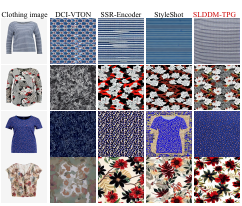}
\caption{Generalization performance comparisons of different methods on the \textit{VITON-HD} dataset \textit{without GT}.}
\label{fig:viton}
\end{figure}

For generalization testing, we randomly sampled 1,000 images from VITON-HD and chose a representative strong baseline from each of three distinct domains for comparisons. As shown in Table~\ref{tab:vtion}, our method achieved the best performance in both visual similarity (VLS) and content similarity (CTS). Figure \ref{fig:viton} presents qualitative comparisons, showing that our model effectively captures diverse clothing content and structured pattern information. The fine-grained details in the generated pattern images align closely with the input clothing images, unlike baseline methods that overlook feature confusion, resulting in visual artifacts. 
\begin{table}[!h]
\renewcommand\arraystretch{0.98}
\small
\centering
\setlength{\tabcolsep}{2.05mm}{
\begin{tabular}{c *4{c}}
  \toprule
\textbf{\small Method} & \textbf{\small FID$\downarrow$} & \textbf{\small SSIM$\uparrow$} & \textbf{\small LPIPS$\downarrow$} & \textbf{\small FPS$\uparrow$} \\

\midrule
{\small DCI-VTON} & 36.39 & 0.175  & 0.521 & 0.758 \\
{\small Paint-by-Example} & 37.52 & 0.167 & 0.536 & 0.746 \\
{\small StrDiffusion} & 39.31 & 0.164 & 0.529 & 0.733 \\
\midrule
{\small SSR-Encoder} & \underline{19.69} & \underline{0.278} & \underline{0.425} & 0.770 \\
{\small IP-Adapter} & 19.98 & 0.272 & 0.431 & \underline{0.773} \\
{\small UniCon} & {21.98} & 0.270 & 0.431 & 0.768 \\
\midrule
{\small StyleShot} & 39.76 & 0.149 & 0.597 & 0.681 \\
{\small OSASIS} & 42.36 & 0.140 & 0.619 & 0.651 \\
\midrule
{\small \bf SLDDM-TPG (Ours)} & \textbf{15.59} & \textbf{0.394} & \textbf{0.280} & \textbf{0.875} \\
{\small \it{improvement}} & \it{4.10} & \it{0.116} & \it{0.145} & \it{0.102} \\
\bottomrule
\end{tabular}}
\caption{Performance comparison for textile pattern generaion on our \textit{CTP-HD} dataset \textit{with Ground Truth ({GT})}.}
\label{tab:ctp}

\end{table}

\begin{table}[!h]
\renewcommand\arraystretch{0.97}
\small
\centering
\setlength{\tabcolsep}{1.25mm}{
\begin{tabular}{c cc c}  
 \toprule
\textbf{\small Method} & \textbf{\small LPIPS (VLS)$\downarrow$}  & \textbf{\small MSE (VLS)$\downarrow$} & \textbf{\small CTS$\uparrow$} \\
\midrule
{\small DCI-VTON} & 0.621 & 0.139 & 0.554 \\
{\small SSR-Encoder} & \underline{0.513} & \underline{0.115} & \underline{0.641} \\
{\small StyleShot} & 0.597 & 0.158 & 0.615 \\
\midrule
{\small \bf SLDDM-TPG (Ours)} & \textbf{0.406} & \textbf{0.080} & \textbf{0.736} \\
{\small \it{improvement}} & \it{0.107} & \it{0.035} & \it{0.095} \\
\bottomrule
\end{tabular}}
\caption{Comparisons of model generalization performance on the \textit{VITON-HD} dataset \textit{without GT}.}
\label{tab:vtion}
\end{table}

\subsection{Ablation Study}
\begin{figure*}[t]
    \centering
\includegraphics[width=0.99\textwidth, trim=0 0 0 3, clip]{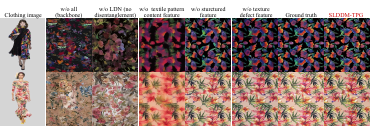}
\caption{Qualitative ablation study on LDN using each individual feature of LDN and the original undisentangled feature.}
\label{fig:ablation_ldn}
\end{figure*}

\noindent\textbf{Multi-dimensional Features in LDN.} We conduct an ablation study to evaluate the effect of each component in our LDN for resolving feature confusion. As shown in Table~\ref{tab:ablation}, removing LDN, especially the textile pattern content feature $f^c_S$ from SCM, model performance drops significantly, since this feature captures key semantic information crucial for reconstruction. The texture defect feature $f^c_T$ from RAM and the structured feature $f^c_A$ from SATs further refine the output by helping the model avoid defects and recover a well-structured layout. When all features are used together, the model achieves the best performance. Qualitative results in Figure~\ref{fig:ablation_ldn} support this. Without disentanglement, derectly using the entangled features lead to blurry, distorted outputs, while excluding structure or defect features causes deformation of visual elements, such as distorted floral shapes, blurry texture, and color degradation. In contrast, incorporating all the disentangled features from LDN yields pattern images with finer granularity, better structure, and higher fidelity.

\begin{table}[!t]
\renewcommand\arraystretch{0.7}
\small
\centering
\setlength{\tabcolsep}{0.45mm}{
\begin{tabular}{c c *4{c}}
  \toprule
\textbf{\small Component} & \textbf{\small Method} & \textbf{\small FID$\downarrow$} & \textbf{\small SSIM$\uparrow$} & \textbf{\small LPIPS$\downarrow$} & \textbf{\small FPS$\uparrow$} \\
\midrule
\multirow{5}{*}{LDN} 
& {\small w/o all} & 41.53 & 0.122 & 0.623 & 0.621 \\
& {\small w/o LDN} & 27.15 & 0.183 & 0.479 & 0.703 \\
& {\small w/o content feature} & 23.59 & 0.219 & 0.468 & 0.719 \\
& {\small w/o structured feature} & 16.04 & 0.376 & 0.315 & 0.863 \\
& {\small w/o defect feature} & 16.89 & 0.376 & 0.320 & 0.857 \\
\midrule
\multirow{3}{*}{S-LDM} 
& {\small w/o alignment process} & 18.61 & 0.350 & 0.318 & 0.852 \\
& {\small w/o CLS module} & 16.23 & 0.364 & 0.299 & 0.727 \\
& {\small w/o STD loss} & 16.86 & 0.359 & 0.311 & 0.863 \\
\midrule
& {\small \bf SLDDM-TPG (Ours)} & \textbf{15.59} & \textbf{0.394} & \textbf{0.280} & \textbf{0.875} \\
\bottomrule
\end{tabular}}
\caption{Ablation study on the Latent Disentangled Network (LDN) and the alignment process in S-LDM.}
\label{tab:ablation}
\end{table}

\begin{figure}[!t]
\centering
    \includegraphics[width=0.48\textwidth, trim=0 0 0 2, clip]{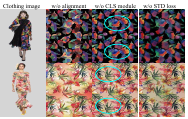}
\caption{Qualitative ablation study on alignment process.}
\label{fig:ablation_sldm}
\end{figure}

\begin{figure}[t]
\centering
\includegraphics[width=0.48\textwidth, trim=0 1 0 0, clip]{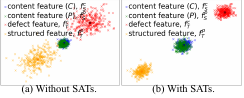}
\caption{The visualization of the distribution movement of LDN's features with and without SATs.}
\label{fig:sat}
\end{figure}

\noindent\textbf{Feature Distribution Movement Analysis.} To evaluate the effect of co-training RAM with SATs, we visualize the LDN's feature distributions with and without SATs. We quantify the distributional difference using the Euclidean distance. As shown in Figure \ref{fig:sat}, without SATs, the texture defect and structured features are separated but fail to converge, indicating drift from the latent space and loss of semantics. In contrast, with SATs, the affine transformations encourage the two features towards convergence, verifying that SATs co-training ensures proper disentanglement.

\noindent\textbf{Ablation on Alignment Process.} Table~\ref{tab:ablation} shows that the CLS module enhances content periodicity (local similarity) of the generated results, as quantified by FPS. Meanwhile, the two ellipses in Figure~\ref{fig:ablation_sldm} highlight shape inconsistencies in regions that were expected to be locally aligned, due to the lack of the CLS module. Removing the STD loss results in fidelity degradation (e.g., color darkening), as the model loses guidance from real paired-data transformations to supervise the mapping from clothing images to generated pattern images. In contrast, incorporating STD loss preserves alignment and consistently improves generation quality. Additional experiments (user study and model efficiency test), extended visualization results, the limitation of our work and future work are provided in the Appendix.

\section{Conclusion}
We propose textile pattern generation (TPG), a new task that synthesizes textile pattern images based on clothing images, and introduce SLDDM-TPG as an effective solution. SLDDM-TPG combines a latent disentangled network (LDN) to resovle feature confusion and a semi-supervised latent diffusion model (S-LDM) to leverage both unlabeled and limited labeled data. This framework enables high-fidelity pattern generation while preserving faithful clothing details. Extensive experiments on our CTP-HD dataset and the widely used VITON-HD dataset confirm the superior performance and good generalization of our approach.

\section{Acknowledgments}
This work was sponsored by National Natural Science Foundation of China (62576305), Zhejiang Provincial Natural Science Foundation of China (LQ24F020020, LD24F020011), ``Pioneer'' and ``Leading Goose'' R\&D Program of Zhejiang (2024C01167), and the Fundamental Research Funds for the Central Universities (No. 226-2025-00057).

\bibliography{aaai2026}

\end{document}